\documentclass{mva_style}
\usepackage{graphicx}

\finalcopy %Uncomment this line for the Camera-Ready Manuscript

\begin{document}
\title{Small Bird Detection using YOLOv7 with Test-Time Augmentation }

\author{
 Kosuke SHIGEMATSU\\
 National Institute of Technology, Oita College\\
 Department of Information Engineering\\
 1666 Maki, Oita, Japan\\
%  {\tt k-shigematsu@oita-ct.ac.jp}\\
}

\maketitle

\section*{\centering Abstract}
\textit{
In this paper, we propose a method specifically aimed at improving small bird detection for the Small Object Detection Challenge for Spotting Birds 2023. Utilizing YOLOv7 model with test-time augmentation, our approach involves increasing the input resolution, incorporating multiscale inference, considering flipped images during the inference process, and employing weighted boxes fusion to merge detection results. We rigorously explore the impact of each technique on detection performance. Experimental results demonstrate significant improvements in detection accuracy. Our method achieved a top score in the Development category, with a public AP of 0.732 and a private AP of 27.2, both at IoU=0.5.
}

\section{Introduction}
The Small Object Detection Challenge for Spotting Birds, held in conjunction with MVA2023\cite{MVA,MVA-proc}, focuses on the Small Object Detection (SOD) problem, which has recently gained attention in the computer vision community due to its unique challenges and potential real-world applications.

In this paper, we propose a method to improve small bird detection performance in object detection competitions using YOLOv7 \cite{yolov7} with test-time augmentation. Our approach involves increasing the input resolution, incorporating multiscale inference, considering flipped images during the inference process, and employing weighted boxes fusion\cite{wbf} to merge detection results. We explore the impact of each technique on the detection performance and demonstrate that our method achieves significant improvements in detection accuracy.

\section{Proposed Method}
\begin{figure}[t]
 \centering
 \includegraphics[width=\linewidth]{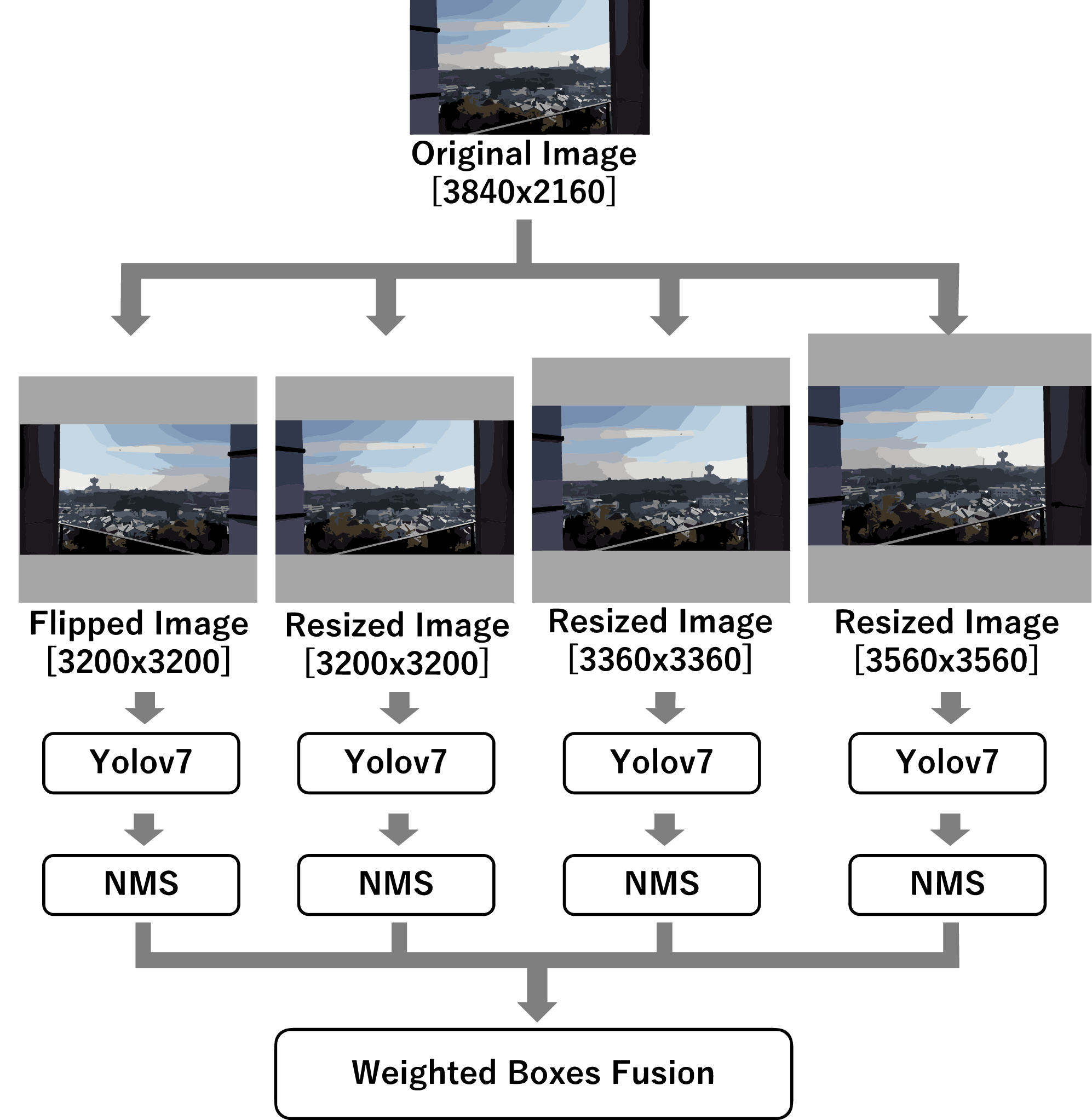}
 \caption{Overview of the proposed algorithm}
 \label{fig:algorithm_overview}
\end{figure}

Our method to improve small bird detection performance using YOLOv7 with test-time augmentation is illustrated in Figure \ref{fig:algorithm_overview}. According to a study in \cite{tta}, the performance of object detection models can be improved by augmenting the data and ensembling the results during testing. Inspired by this idea, we seek to improve model performance by combining results from flipping and resizing images during inference. The method consists of three key components: increasing the input resolution, incorporating multiscale inference and flipped image consideration, and employing weighted boxes fusion to merge detection results.

\subsection{Increasing the Input Resolution}
To enhance the detection accuracy of the YOLOv7 model, we increased the input resolution of the model. The model was tested with different resolutions, including 1280×1280, 2560×2560, and 3200×3200.

\subsection{Multiscale Inference and Flipped Image Inference}
We incorporated multiscale inference by performing object detection at different input resolutions, specifically 3200×3200, 3360×3360, and 3520×3520. Additionally, flipped images were considered during the inference process. The detection results from the multiscale images and the flipped images were then merged using weighted boxes fusion.

% \subsection{Weighted Boxes Fusion}
% Weighted boxes fusion was employed to merge the detection results obtained from the multiscale inference and flipped image consideration. This technique proved effective in improving the detection accuracy by combining the predictions from different sources and leveraging their complementary strengths.

\section{Experiments and Results}
In this section, we present the experimental setup and the results obtained from the proposed method, showcasing the effectiveness of the YOLOv7 with test-time augmentation in detecting small birds. 

\subsection{Experimental Setup}
The experiments were conducted using the YOLOv7 model with different input resolutions, incorporating multiscale inference, flipped image consideration, and data augmentation. The dataset used for training consists of 9,759 images with 29,037 annotated bird instances, and we utilized the Train2 (mva2023$\_$sod4bird$\_$train) subset only. The training set was divided into 8,879 images for training and 880 images for validation. We used the initial weights from the YOLOv7 model (yolov7$\_$training.pt) and trained for 1,000 epochs. Data augmentation settings were adjusted, including changes to the degrees (from 0.0 to 5.0) and shear (from 0.0 to 1.0). The performance of the model was evaluated using the Average Precision (AP) metric on the validation set.

\subsection{Results}

Table \ref{table:results} summarizes the experimental results obtained from the proposed method. 
The results demonstrate that increasing the input resolution significantly contributes to improved detection performance. Furthermore, the integration of multiscale inference and flipped image consideration, combined with weighted boxes fusion, led to a marked improvement in detection accuracy. Our method achieved a public AP of 73.2 at IoU=0.5 and a private AP of 27.2 at IoU=0.5 in the Small Object Detection Challenge for Spotting Birds 2023.

\begin{table}[h]
%  \caption{Experimental results of the proposed method}
 \caption{Experimental results of the proposed method (AP with IoU threshold 0.5)}
\label{table:results}
 \centering
  \begin{tabular}{clll}
   \hline
   Input Resolution & Inference &Public & Private \\
                    & Strategy  &AP     & AP \\
   \hline \hline
   1280×1280 & -                & 49.4 & -    \\
   2560×2560 & -                & 65.7 & -    \\
   3200×3200 & -                & 70.5 & -    \\
   3200×3200 & Multiscale       & 73.1 & -    \\
   3200×3200 & Multiscale \&    & 73.2 & 27.2 \\
             & Flipped          &      &      \\
   \hline
  \end{tabular}
\end{table}

\section{Conclusion}

In this paper, we presented a novel, lightweight approach for small bird detection using YOLOv7 with test-time augmentation. Our method focused on increasing the input resolution and employing multiscale and flipped image inferences, followed by merging the results using weighted boxes fusion. The experimental results demonstrated significant improvements in the mean Average Precision (AP) compared to the original YOLOv7, with the additional benefit of requiring relatively fewer parameters.

Our approach achieved notable success in the Small Object Detection Challenge for Spotting Birds 2023 Development category, attaining a top score with a public AP of 0.732 and a private AP of 27.2, both at IoU=0.5. These results confirm the effectiveness of the proposed method for small bird detection in challenging scenarios and validate its potential applicability in real-world settings such as avoiding bird attacks and driving away harmful birds in agricultural environments.

In future work, we plan to explore additional enhancements and optimizations to further improve the performance and applicability of our method in various small object detection tasks.


\begin{thebibliography}{99}

\bibitem{MVA}
  Small Object Detection Challenge for Spotting Birds 2023: {\tt https://www.mva-org.jp/mva2023/challenge}.

\bibitem{MVA-proc}
 Yuki Kondo, et al.: 
  ``MVA2023 Small Object Detection Challenge for Spotting Birds: Dataset, Methods and Results,'' 
  \textit{2023 18th International Conference on Machine Vision and Applications (MVA) Workshop}, 2023.

\bibitem{yolov7}
  Chien-Yao Wang, et al.: 
  ``YOLOv7: Trainable bag-of-freebies sets new state-of-the-art for real-time object detectors,'' 
  \textit{arXiv},  arXiv:2207.02696, 2022.

\bibitem{wbf}
  Roman Solovyev , et al.: 
  ``Weighted boxes fusion: Ensembling boxes from different object detection models,'' 
  \textit{Image and Vision Computing}, vol.107, p.104-117, 2021.

\bibitem{tta}
  Casado-García, Á., et al.: 
  ``Ensemble methods for object detection,'' 
  \textit{In Proceedings of the ECAI},  2020.


\end{thebibliography}
\end{document}